\pdfoutput=1

\documentclass[11pt]{article}

\usepackage[final]{acl}

\usepackage{times}
\usepackage{latexsym}

\usepackage[T1]{fontenc}

\usepackage[utf8]{inputenc}

\usepackage{microtype}

\usepackage{inconsolata}

\usepackage{hyperref}
\usepackage{url}
\usepackage{graphicx}
\usepackage{xspace}
\usepackage{algorithm}
\usepackage{algorithmic}

\providecommand{\ourModel}{\textsf{PLHF}\xspace}

\title{PLHF: Prompt Optimization with Few-Shot Human Feedback}

\author{Chun-Pai Yang$^\dag$, Kan Zheng$^\ddag$, and Shou-De Lin$^\ast$ \\
  Intelli-Train.ai$^{\dag\ddag\ast}$, ZRT Technology$^{\ddag}$, National Taiwan University$^{\dag\ast}$ \\
  \texttt{chunpai@intelli-train.ai}, \texttt{kanzheng05@intelli-train.ai}, \texttt{sdlin@csie.ntu.edu.tw} \\}

\begin{document}
\maketitle
\begin{abstract}
Automatic prompt optimization frameworks are developed to obtain suitable prompts for large language models (LLMs) with respect to desired output quality metrics. Although existing approaches can handle conventional tasks such as fixed-solution question answering, defining the metric becomes complicated when the output quality cannot be easily assessed by comparisons with standard golden samples. Consequently, optimizing the prompts effectively and efficiently without a clear metric becomes a critical challenge. To address the issue, we present \ourModel (which stands for \textbf{P}rompt \textbf{L}earning with \textbf{H}uman \textbf{F}eedback), a few-shot prompt optimization framework inspired by the well-known RLHF technique. Different from n\"aive strategies, \ourModel employs a specific evaluator module acting as the metric to estimate the output quality. \ourModel requires only a single round of human feedback to complete the entire prompt optimization process. Empirical results on both public and industrial datasets show that \ourModel outperforms prior output grading strategies for LLM prompt optimizations.
\end{abstract}

\section{Introduction}
\label{section:introduction}

General-purpose large language models (LLMs) have shown substantial capabilities across various fields in recent years. However, solving complex tasks with LLMs often requires appropriate customizations on LLMs to fit the task requirements. While fine-tuning pre-trained LLMs is a common approach, it may be infeasible when there is limited training data, restricted computational resource, or when working with a black-box LLM. Alternatively, previous studies~\citep{wang2022no, shin2020autoprompt} have shown that the potential of LLMs can also be fully leveraged with suitable \emph{prompts}.
Recent literature develops automatic few-shot prompt optimization for LLM usages, such as \textsf{DSPy}~\citep{khattab2024dspy} and \textsf{TextGrad}~\citep{textgrad}. To determine an effective prompt for the LLM, existing methods often employ gradient descent or other algorithms~\citep{yang2024large,zhou2023largelanguagemodelshumanlevel,guo2023connecting} to optimize the performance with respect to desired metrics (i.e., definition of output quality). Overall, the key to success heavily relies on the output quality evaluations which shall precisely reveal the model performance to the optimizer.

\begin{figure*}[t]
\centering
\includegraphics[width=\linewidth]{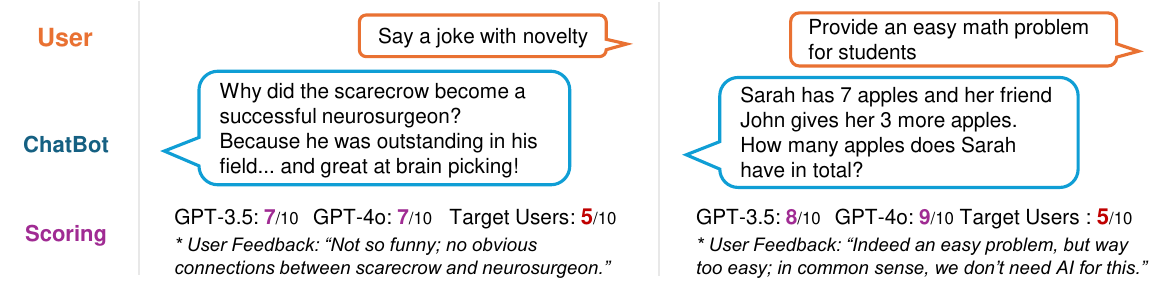}
\caption{Demonstrations of the actual failure cases that the evaluations from pre-trained LLMs have different preference from specific humans. The first (left) example is the task of joke generation, where the grading is according to funniness and novelty. The second scenario is a math problem generation bot, where the response quality is evaluated based on  helpfulness and problem quality. As shown above, the verdicts of state-of-the-art LLMs could still differ from real human's preferences.}
\label{fig:failure-examples}
\end{figure*}

\begin{figure*}[t]
  \centering
  \includegraphics[width=\linewidth]{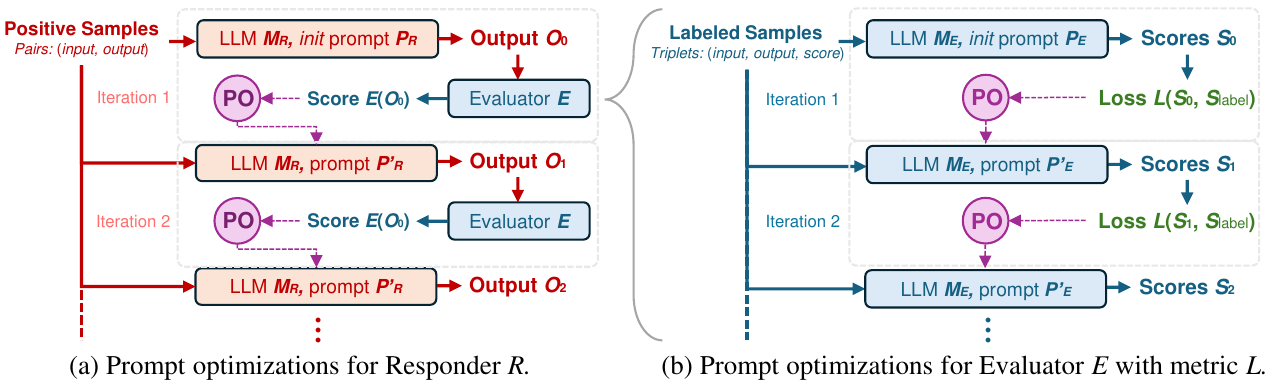}
  \caption{Workflow framework of \ourModel. The entire LLM program contains two modules, Responder $R$ and Evaluator $E$, where PO can be PO arbitrary prompt optimization method.}
  \label{fig:framework-simple}
\end{figure*}

Although such output quality metrics are often well-defined for the tasks which can be modeled as the traditional discriminative tasks (e.g., classifications and regressions) where the performance can be directly evaluated given ground-truths, grading the outputs often becomes non-trivial for most generation-type of tasks.
The absence of a precise metric would hinder the effectiveness of the prompt optimization process for a generative task.

Most prompt optimization methods adopt two types of mechanisms to evaluate the generated outputs. The first is employing simple evaluators, such as exact-matching or soft-matching (based on similarity) to compare the generated outputs with the observed samples. The second common strategy is to utilize pre-trained LLMs for the output grading. For instance, several studies~\citep{wang2023chatgpt,fu2024gptscore} leverage advanced OpenAI GPT models~\citep{achiam2023gpt} as the judges to grade the generated outputs. Nevertheless, such graders suffer from a critical drawback as generic pre-trained LLMs might not have enough contextual or background knowledge to behave as accurate evaluators. (See Figure~\ref{fig:failure-examples} for the examples of failure cases.) As a result, it is highly demanded to involve real human experts to evaluate generated results from LLMs, yet such schemes often suffer from budget constraints of employing human experts.

For most of the automatic prompt optimization frameworks~\citep{khattab2024dspy,textgrad,pryzant2023automatic,deng2022rlprompt,wen2024hard}, multiple iterations are performed with a given output quality metric. With human expert inquiries acting as the metric, once the prompt is updated with any modifications, we have to ask human experts for their judgment repeatedly for each sample --- this causes serious efficiency bottleneck. To address the aforementioned issues, we present \ourModel, a novel few-shot prompt optimization framework with \textit{two} specific modules. Inspired by the famous Reinforcement Learning from Human Feedback (RLHF) technique~\citep{ouyang2022training,li2023reinforcement}, \ourModel introduces a particular \emph{evaluator} module that requires human grading no greater than \emph{linear} (with respect to the number of training samples) times during the optimization process. To leverage human feedback in few shots, we consider utilizing a prompt-optimized LLM as $E$ to evaluate the output of the main responder $R$. The overall framework is depicted in Figure~\ref{fig:framework-simple}.

Specifically speaking, first, we employ human experts to provide judgments as scores on a set of training samples, containing input-output pairs. Then, we perform prompt optimization on another LLM to mimic the human experts' preference pattern. Since the prompting task on the evaluator module is relatively typical (e.g., binary classifications or regressions), we can leverage any existing automatic few-shot prompt optimization frameworks (e.g., \textsf{DSPy}~\citep{khattab2024dspy}) with trivial metrics (e.g., Accuracy or Mean Absolute Error) to obtain the evaluator module. Finally, we can take advantage on a prompt-optimized evaluator acting as the metric for the original task.

To verify the actual effectiveness of \ourModel, we conduct experiments on multiple public datasets and a real-world industrial data collected from a customer support chat-bot product of an AI company. The experimental results shown in Section \ref{section:experiments} demonstrate that \ourModel can boost the output quality of existing automatic few-shot prompt optimization frameworks with our duo-module design.

In summary, our contributions are as follows:
\begin{itemize}
    \item We study few-shot prompt optimization for LLMs with limited number of human feedback calls, which is a more reasonable and feasible setting in real-world applications.
    \item We introduce \ourModel, a novel prompt optimization framework that does not directly rely on well-defined evaluation metrics. Instead, we design an evaluator module to provide an automatic grading mechanism.
    \item With extensive experiments on public datasets and an industrial dataset, we show that \ourModel has superiority toward output quality, compared with the approaches employing string matching or adopting the state-of-the-art LLMs (e.g., \textsf{GPT-4o}) as the evaluator.
\end{itemize}

We have also surveyed relevant prior efforts, which are summarized into a Related Work section in Appendix A.

\section{PLHF: Prompt Optimization with Few-shot Human Feedback}
\label{section:methodology}

\begin{algorithm*}[h]
\caption{\ourModel: A duo-module Framework for Few-shot LLM Prompt Optimizations}
\label{alg:framework}

\begin{algorithmic}[1]
\REQUIRE Training data samples $D = \{d_1, d_2, \dots, d_n\}$ (each $d_i$ is a triplet of input/query $q_i$, output $o_i$ and score/verdict $r_i$), Base LLMs $M_R$, $M_E$, Initial prompts $P_R, P_E$, The trivial metric $L$ for evaluator $E$.
\ENSURE Optimized prompts $P'_R$ and $P'_E$.

\STATE Set $P'_R := P_R$ and $P'_E := P_E$.
\WHILE{there are new training samples added into $D$}
    \WHILE{$P'_E$ is not yet optimized (converged) for $D$}
        \FOR{each training sample $d_i = (q_i, o_i, r_i) \in D$}
            \STATE Input $(q_i, o_i)$ pair to generate the score (or verdict) $\tilde{r}_i$ by $M_E$ with prompt $P'_E$.
        \ENDFOR
        \STATE Compute conventional metric score $S_E = L(\{\tilde{r}_1, \dots, \tilde{r}_n\}, \{r_1, \dots, r_n\})$.
        \STATE Consider the $D$, $P'_E$ and $S_E$ to update the prompt $P'_E$.
    \ENDWHILE
    \WHILE{$P'_R$ is not yet optimized (converged) for $D$}
        \FOR{each training sample $d_i = (q_i, o_i, r_i) \in D$ with \emph{positive} rating $r_i$}
            \STATE Input $q_i$ to generate the output $\tilde{o_i}$ by $M_R$ with prompt $P'_R$.
        \ENDFOR
        \STATE Evaluate the outputs by evaluator $E$. Obtain the score $S_R = M_E(\{\tilde{o}_1, \dots, \tilde{o}_n\}; P'_E)$.
        \STATE Consider the $D$, $P'_R$ and $S_R$ to update the prompt $P'_R$.
    \ENDWHILE
\ENDWHILE

\RETURN $P'_R, P'_E$
\end{algorithmic}
\end{algorithm*}

As shown in Figure~\ref{fig:framework-simple}, our entire framework, \ourModel, is designed to perform prompt optimizations for typical language model program usages --- output a proper response based on the given input. The whole process is guided by the principal intuition of few-shot in-context learning \cite{brown2020language} to capture contextual patterns from a limited number of labeled samples. Since there is \textit{no} explicit metric available, a grading function is needed for existing prompt optimization methods. Hence, we introduce an \emph{evaluator} module $E$, acting as the grading function for the main \emph{responder} module $R$. The two modules correspond to two respective subtasks.

In Appendix B, we provide an example of a toy problem to demonstrate the initial prompts and the optimized prompts obtained from \ourModel.

\subsection{Evaluator Task}

The auxiliary task for evaluator $E$ is designed to \textit{grade} the outputs generated by the responder $R$. The evaluator $E$ is built with a base LLM, denoted as $M_E$. The training samples for $E$ are triplets of input, output, and the corresponding score graded by human experts. To provide a nuanced verdict, we also optimize the prompt $P_E$ for $M_E$ with a \textit{trivial} metric $L$ (e.g., the conventional Accuracy or Mean Absolute Error) to evaluate the quality of the response. The metric $L$ is specifically defined as the loss to estimate the difference between a predicted score and the actual graded score.

With the grading ability of the evaluator $E$, \ourModel ensures that the outputs from the responder $R$ are not only technically accurate but also contextually appropriate for the task. Most importantly, the entire evaluation process operates without any human labor expenses.

\subsection{Responder Task}

The task for responder $R$ is performing the original assignment. The core of this component is also a base LLM, denoted as $M_R$, which generates outputs for input queries. For $M_R$, the prompt is starting with an initial setting $P_R$ which simply describes the task. The training samples for $R$ include input-output pairs with \emph{positive} score/verdict labeled by human experts. The data positivity can be specifically defined to align with the requirements of the assigned AI task. The generated responses by the LLM $M_R$ with prompt $P_R$ are then judged by our evaluator module $E$ to perform prompt optimizations on LLM $M_R$ to obtain a \textit{prompt-optimized} prompt $P'_R$ for $M_R$.

\subsection{Duo-module Integration}

With the integration of responder $R$ and evaluator $E$, the entire system operates as described in Algorithm~\ref{alg:framework}. At the beginning, we initialize prompts as $P'_R := P_R$ and $P'_E := P_E$ for modules $R$ and $E$, respectively. In each iteration of \ourModel, first, training data samples $D$ are used to optimize the evaluator $E$ (i.e., to update $P'_E$). Then, we optimize the responder $R$ (i.e., update $P'_R$) regarding the evaluator $E$ with prompt $P'_E$ as the metric. After an iteration of prompt optimizations for $P'_R$ and $P'_E$, we obtain a version of optimized prompts for the responder $R$ and the evaluator $E$, respectively.


Overall, the proposed \ourModel framework is capable of performing prompt optimizations for LLMs even when occurring challenges of (a) no available well-defined metrics to evaluate the LLM output quality for the specific task, (b) limited number (few-shot) of labeled samples for LLM prompting, and (c) multiple valid outputs for a single input.

\section{Experiments}
\label{section:experiments}

To evaluate the performance and robustness of our proposed model framework, \ourModel, we conducted a series of experiments across various tasks. The tasks were selected to test the model ability to generate accurate and contextually relevant outputs, while also assessing the effectiveness of the auxiliary evaluator task in refining responses.

\subsection{Datasets}

We conduct the experiments on three public datasets and one industrial dataset from a real-world scenario of question answering chat-bot generating SQL commands.

\subsubsection{Schema Guided Dialogue Dataset}

The Schema Guided Dialogue (SGD) dataset~\cite{dataset-sgd} is a large-scale dataset designed for task-oriented dialogue systems. It comprises dialogues collected in English, specifically designed to encompass a wide range of dialogue scenarios, schema-based actions, and services. The dataset contains 1,000 dialogues, contributing to a total of 13,833 utterances. Each user utterance in the dataset is labeled with a \emph{satisfaction score} on a 5-point Likert scale~\citep{likert1932technique}. Rating 1 indicates the lowest level of satisfaction, while rating 5 denotes the highest satisfaction. The human-assigned satisfaction scores are valuable for assessing chat-bot responses with respect to user satisfaction.

\subsubsection{Automated Essay Scoring Datasets}

The dataset is originally provided by \cite{asap-aes} for the Automated Student Assessment Prize (ASAP). The dataset, named as AES-ASAP, consists of eight essay sets varying in length, ranging from an average of 150 to 550 words per response. The responses were written by students in grades seven through ten, and all essays were hand-graded by human experts. Each essay was double-scored. We use the training set of the first essay set for our experiments. The actual text of the student's response is included. We consider \emph{average score} as the aggregated result from the raters. The scores are distributed from 1 to 30. In our experiments, we discard the essays with transcription errors (marked as ``illegible'' or containing placeholder text such as ``???'') from the training data.

Apart from AES-ASAP, our experiments also include a newer essay scoring dataset of from an online competition hosted by \cite{aes-dataset-2}. The dataset, named as AES-2.0, contains 24,000 student-written argumentative essays. Each essay was scored on a scale of 1 to 6 as the holistic rating \footnote{\url{https://storage.googleapis.com/kaggle-forum-message-attachments/2733927/20538/Rubric_ Holistic Essay Scoring.pdf}} judged by human experts.

\subsubsection{Industrial SQL Command Question Answering Dataset}

In addition to the previously mentioned public datasets, we have also deployed \ourModel on a real-world question-answering system, which is currently an actual product of a commercial AI company. This test, named as SQL-QA, comprises 100 real-world queries involving various database inquiry requests from the clients. The database entries contains daily transaction records and other logs sourced from multiple banks in China. As the training data for prompt optimizations, human experts from the company labeled 10 positive samples and 20 negative samples for the prompt optimizations. In Appendix C, we provide some examples of queries used in this test.

\subsection{Experiment Setups}

\begin{table*}[t]
\centering
\caption{Summary of experimental results for the evaluator subtask across each dataset. For the public datasets, the presented values are RMSE losses (lower is better) of the output scores from $E$; for the industrial dataset {SQL-QA}, the values indicate Accuracy scores (higher is better). The best ones are marked in bold font.}

\begin{tabular}{c|c|c|c|c}
\textbf{Method} & \ \ \ \textbf{SGD}\ \ \  & \textbf{AES-ASAP} & \textbf{AES-2.0} & \textbf{SQL-QA} \\
\hline
Base LLM (\textsf{GPT-3.5}) & 1.02 & 4.75 & 0.46 & 0.53 \\
\hline
MLP with Text Embedding & 1.17 & 7.22 & 1.08 & 0.33 \\
SVM with Text Embedding & 1.25 & 6.43 & 1.10 & 0.40 \\
\hline
Base LLM PO via \textsf{DSPy} & 0.43 & \textbf{2.36} & \textbf{0.33} 
& \textbf{0.80} \\
Base LLM PO via \textsf{TextGrad} & \textbf{0.40} & 2.42 & 0.38 & 0.73 \\
\end{tabular}
\label{table:evaluator-results}
\end{table*}

\begin{table*}[t]
\centering
\caption{Summary of experimental results for the responder subtask across each dataset. For the industrial dataset SQL-QA, the overall Accuracy score are given by actual human experts in the company; for the public datasets, the scores from the pseudo-human judge are shown. The values for Base LLM are the actual scores, whereas for the other methods, relative improvements are shown in percentages. The best ones are marked in bold font.}

\begin{tabular}{c|c|c|c|c|c}
\textbf{PO} & \textbf{Method} & \ \ \ \textbf{SGD}\ \ \  & \textbf{AES-ASAP} & \textbf{AES-2.0} & \textbf{SQL-QA} \\
\hline
~ & Base LLM (\textsf{GPT-3.5}) & 4.25 & 26.50 & 5.35 & 0.74 \\
\hline
~ & PO with \textsf{GPT-4o} & +1.18\% & +3.70\% & +0.75\% & +5.41\% \\
\textsf{DSPy} & PO with Exact Matching & - 4.00\% & -10.87\% & - 5.61\% & 0.00\% \\
~ & PO with Embedding Similarity & +1.65\% & - 4.27\% & +0.56\% & +10.81\% \\
~ & \ourModel & \textbf{+6.59\%} & \textbf{+8.45\%} & \textbf{+2.62\%} & \textbf{+18.92\%} \\
\hline
~ & PO with \textsf{GPT-4o} & +4.71\% & +3.28\% & +1.31\% & +2.70\% \\
~ & PO with Exact Matching & -10.59\% & -15.92\% & -10.84\% & -18.92\% \\
\textsf{TextGrad} & PO with Embedding Similarity & +3.53\% & - 0.64\% & +0.93\% & +2.70\% \\
~ & \ourModel & \textbf{+8.71\%} & \textbf{+8.68\%} & \textbf{+4.30\%} & \textbf{+18.92\%} \\
\end{tabular}
\label{table:responder-results}
\end{table*}

To perform prompt optimizations (denoted as \textbf{PO}) in each subtask, we consider two state-of-the-art automatic prompting frameworks, \textsf{DSPy} and \textsf{TextGrad}. Same experiments are conducted for both frameworks, and respective results are shown.

For the experiments, first, we estimate the effectiveness of the evaluator $E$, which solves the task of predicting the labeled scores based on each input-output pair. The comparisons include several baseline methods:

\begin{itemize}
    \item \textbf{Base LLM (\textsf{GPT-3.5})}: the grounding baseline --- simply employ OpenAI \textsf{gpt-3.5-turbo-0125} model to predict the labeled score each input-output pair.
    \item \textbf{MLP with Text Embedding}: Leveraging a 3-layer Perception model, based on~\cite{mlp}, to predict the score of each input-output pair. The texts are transformed into embedding vectors via OpenAI \textsf{text-embedding-ada-002} model~\citep{text-emb-ada-002}.
    \item \textbf{SVM with Text Embedding}: similar to the MLP one, but adopt Support Vector Machines (SVMs) as the score predictor. We adopt the implementation provided by~\cite{libsvm} with the default configuration.
    \item \textbf{\textsf{GPT-3.5} PO via DSPy}: perform prompt optimizations with \textsf{DSPy}~\citep{khattab2024dspy}.
    \item \textbf{\textsf{GPT-3.5} PO via TextGrad}: perform prompt optimizations with \textsf{TextGrad}~\citep{textgrad}.
\end{itemize}

Then, for the main responder $R$, output quality of the LLM with optimized prompts are judged by test queries. We consider various types of the evaluators for prompt optimizations.
\begin{itemize}
    \item \textbf{Base LLM (\textsf{GPT-3.5}, no PO)}: the grounding baseline --- simply utilize OpenAI \textsf{gpt-3.5-turbo-0125} model to generate the output based on the given input.
    \item \textbf{PO with \textsf{GPT-4o}}: employ the state-of-the-art \textsf{GPT-4o} (\textsf{gpt-4o-2024-05-13}) as the evaluator for prompt optimizations on the Base LLM (\textsf{GPT-3.5}).
    \item \textbf{PO with Exact Matching}: consider hard-matching as the grading function for the prompt optimization on the Base LLM. Let score $= 1$ if and only if the output of $R$ is exactly the same as the ground-truth.
    \item \textbf{PO with Embedding Similarity}: similar to the Exact Matching one, but this time consider cosine similarity as the score between embedding vectors. The outputs are embedded by OpenAI \textsf{text-embedding-ada-002} model~\citep{text-emb-ada-002}.
    \item \textbf{\ourModel}: our framework, where LLMs $M_R$ and $M_E$ are both set to be \textsf{GPT-3.5} (\textsf{gpt-3.5-turbo-0125}) for fair comparisons.
\end{itemize}

\subsection{Evaluations}

For the model output, we employ multiple human experts as the judges to provide professional scores with respect to the score scales of the original data. However, for the public datasets, the original people who labeled the data are unavailable to give their judgments for our new generated outputs. Therefore, we introduce a concept of \emph{pseudo-human} judge to perform output evaluations. Specifically, we use \textsf{GPT-4o} with prompt optimizations via \textsf{DSPy} as the pseudo-human judge. Since we consider \textsf{GPT-3.5} as Base LLMs for all the methods in the experiments, this pseudo-human judge is a relatively powerful model that can provide more convincing evaluations toward output quality.

\subsection{Experimental Results}

Table~\ref{table:evaluator-results} lists the experimental results of the evaluator task. As shown in the table, we can observe that the conventional methods (MLP/SVM with Text Embedding) seem struggled in predicting the labeled scores from the given embedded inputs. In contrast, the LLM-based methods performed significantly better on both public datasets and the industrial tests. A possible reason is that LLMs might have superior fitting and understanding capabilities to handle text inputs. With prompt optimizations, both \textsf{DSPy} and \textsf{TextGrad} provided more effective prompt for more accurate evaluators.

As for the experimental results of the responder task, shown in Table~\ref{table:responder-results}, we consider both \textsf{DSPy} and \textsf{TextGrad} as the prompt optimization (PO) tool for each method in the comparisons. Overall, the results are relatively similar in same directions for each pair of the scores between the two PO selections. In summary, for all the four datasets, the proposed \ourModel framework achieved the best performance in output quality, in terms of the metric for each task. Moreover, \ourModel used \textsf{GPT-3.5} as the evaluator's base LLM $M_E$ to achieve superior performance than `PO with \textsf{GPT-4o}', which conducted prompting with a more powerful \textsf{GPT-4o} as the evaluator. For the other baselines, `PO with \textsf{GPT-4o}' consistently outperformed `Base LLM (raw \textsf{GPT-3.5})'. Regarding the conventional matching-based grading functions, both hard-matching (Exact Matching) and soft-matching (Embedding Similarity) produced outputs with worse quality. We have also tested the model performance under various numbers of training samples. The additional results are shown in Appendix D.

\section{Conclusion}
\label{section:conclusion}

In this paper, we focused on LLM prompt optimizations with a limited amount of human feedback --- a more practical and achievable approach in real-world applications. To address the challenges of no well-defined metrics and the scarce human resources, we introduced \ourModel, a few-shot prompt optimization with an evaluator module design to automatically grade the outputs generated by LLMs. We performed extensive experiments with public datasets and a real industrial dataset to verify the effectiveness of \ourModel. The experimental results shown that \ourModel outperforms existing methods across from simple string matching functions to even the latest publicly available LLMs as output evaluators in terms of the output quality. Overall, our proposed framework is practically effective, especially for the scenarios when directly applying pre-trained general-purpose LLMs are not the best option. Our future work involves enhancing and deploying the proposed framework across diverse applications, particularly for tasks that utilize multi-modal data.

\bibliography{custom}

\appendix

\section{Related Work}
\label{section:related-work}

Various strategies have been developed to obtain suitable LLM prompts. Earlier studies adopt automated data sample searching techniques~\citep{gao2021making} to learn prompts through gradient-based searching methods \citep{shin2020autoprompt,wen2024hard,pryzant2023automatic}, refining prompts using evolutionary algorithms \citep{guo2023connecting,fernando2023promptbreeder} and utilizing other LLMs for prompt generation \citep{yang2024large, zhou2023largelanguagemodelshumanlevel}. Several studies have also attempted to optimize prompts using reinforcement learning, exploring prompt editing at different granular levels such as word-level \citep{deng2022rlprompt}, phrase-level \citep{zhang2023tempera}, and within text-to-image generation tasks \citep{hao2024optimizing}.

As LLMs are increasingly applied in real-world scenarios, in-context learning \citep{mccann2018natural,radfordimproving,brown2020language} is becoming an emerging trend for effective LLM programming. Instruction tuning \citep{ouyang2022training} further enhances this process by enabling complex behaviors through the use of structured prompts \citep{press2023measuring,yao2023react,khot2023decomposed,madaan2024self}.

\begin{figure*}[t]
  \centering
  \includegraphics[width=\linewidth]{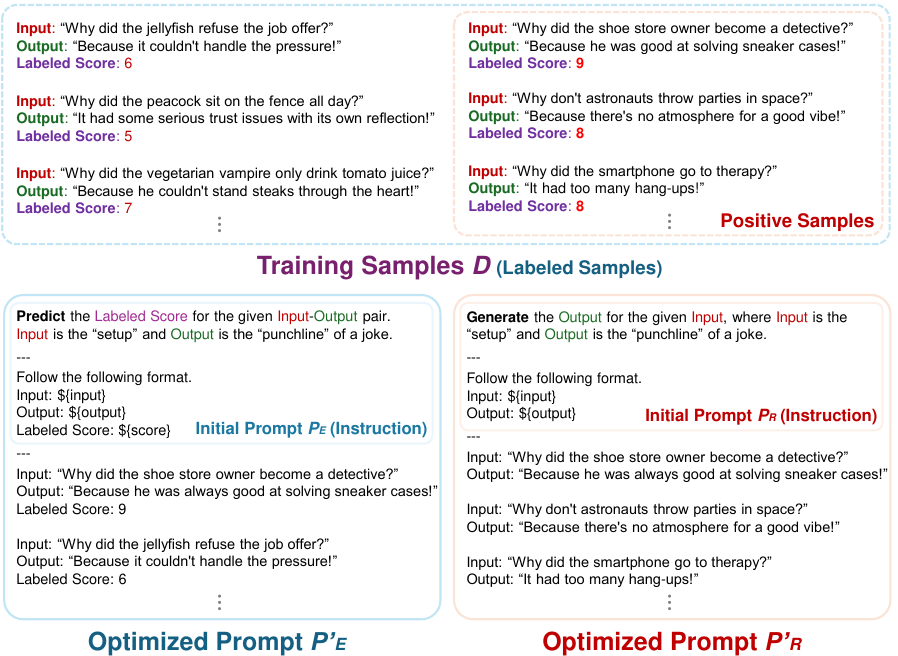} 
  \caption{A toy example to illustrate the subtask designs of \ourModel. The targeted generative AI task for this example is ``\emph{generate the punchline for a joke setup}.'' The training samples $D$ are triplets (\textit{Input}, \textit{Output}, \textit{Labeled Score}), where \textit{Input} is the joke setup, \textit{Output} is a sample output of the punchline for the corresponding \textit{Input}, and \textit{Labeled Score} is the rating judged by human experts. For this example, we consider \textit{Labeled Score} $\geq 8$ as the condition of positive samples. Examples of optimized prompts $P'_E$ and $P'_R$ (for the evaluator $E$ and the responder $R$, respectively) are shown.}
  \label{fig:optimization-example}
\end{figure*}

For automatic few-shot prompt optimization, \cite{khattab2024dspy} introduced \textsf{DSPy}, a state-of-the-art prompt optimization framework, which considers LLM usages in a programmatic fashion. \textsf{DSPy} parameterizes each module to learn the data pattern and the desired behaviors by iteratively bootstrapping useful demonstrations. On the other hand, \cite{textgrad} inspired by LLM fine-tuning procedures and proposed \textsf{TextGrad}, a framework refining the prompt with the back-propagation algorithm. Instead of deriving numeral-valued gradients, \textsf{TextGrad} regards LLMs' text feedback as the `gradient' in texts.

All these techniques rely on well-defined metrics to set their objectives. While advanced general-purpose LLMs like GPT-4 can be adopted for text-output evaluations \citep{eval-llm-with-gpt4,zheng2023judging,tan2024fine}, they may lack the contextual or background knowledge needed for accurate evaluation in specific tasks. Hence, involving human feedback becomes inevitable for the prompt optimizations in such tasks. To address the issue, \cite{lin2024prompt} inspired by dueling bandits and designed a strategy to choose pairs of prompts to query for human feedback during the prompt optimizations to reduce the number of needed calls of human feedback. In this paper, we consider a different approach to tackle the issue --- we focus on the \emph{metric} in the prompt optimization process. We developed a duo-module framework to obtain an evaluator module acting as the metric of the main task to perform the desired LLM prompt optimizations, requiring minimal human feedback.

\section{An Optimization Example of \ourModel}

In Figure~\ref{fig:optimization-example}, we provide an example of a toy problem of automatic joke generation to demonstrate the initial prompts $P_E$ and $P_R$, as well as the optimized prompts $P'_R$ and $P'_E$ obtained from \ourModel.

\begin{table*}[t]
  \centering
  \caption{Examples of the queries in our industrial SQL-QA test.}
  \begin{tabular}{p{5.5cm}|p{8.5cm}}
\textbf{User Query (English translation)} & \textbf{SQL Statement (Output)} \\
\hline
{\footnotesize\begin{verbatim}Please list the top 3
clients by total deposits
at the Beijing branch
as of January 31, 2024.\end{verbatim}} & {\footnotesize\begin{verbatim}SELECT CUST_ID, CUST_NAME, DEPO_BAL
FROM acct
WHERE DATA_DT='20240131' AND ORG_NAME='Beijing'
ORDER BY DEPO_BAL DESC
LIMIT 3\end{verbatim}}
\vspace*{-12pt}\\
\hline
{\footnotesize\begin{verbatim}Please inquire about the
top 5 banks with the
highest asset balances,
grouped by institution,
as of March 31, 2024.\end{verbatim}} & {\footnotesize\begin{verbatim}SELECT ORG_NAME, ASSET_BAL
FROM acct
WHERE DATA_DT='20240331'
ORDER BY ASSET_BAL DESC
LIMIT 5\end{verbatim}}
\end{tabular}
  \label{table:sql-examples}
\end{table*}

\begin{figure*}[!h]
  \centering
  \includegraphics[width=0.875\linewidth]{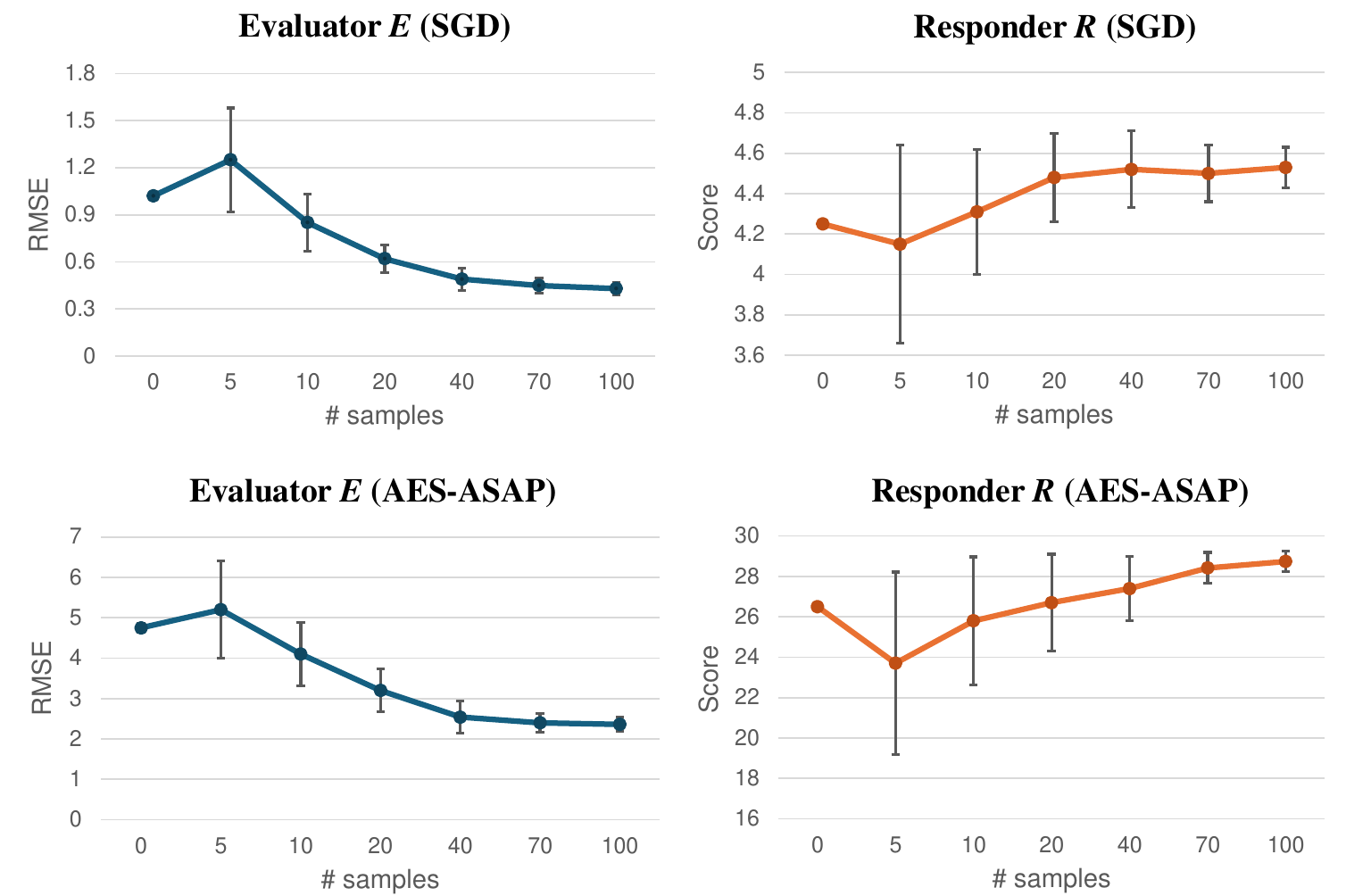} 
  \caption{Performance curves of \ourModel on the datasets SGD and AES-ASAP. For the plots, we consider \textsf{DSPy} as the PO method for \ourModel. The $x$-values are the number of (randomly selected) training samples. The $y$-values are mean values of the RMSE losses for $E$ and the output scores for $R$, respectively. The vertical bar of each point indicates the standard deviations estimated in 30 runs.}
  \label{fig:perf-analysis}
\end{figure*}

\section{Industrial SQL Command Question Answering Dataset}

In Table~\ref{table:sql-examples}, we provide two actual query samples of the industrial SQL-QA dataset which we consider in the experiments.

\section{Performance Analysis with Various Numbers of Training Samples}

In addition to the overall performance, we also analyze the robustness and the relationship between model effectiveness and the number of training samples involved in the prompt optimization process. As examples, Figure \ref{fig:perf-analysis} demonstrates the performance curves for datasets SGD and AES-ASAP. Note that, to analyze the performance of a responder with $n$ data samples, we also use the evaluator optimized with the same $n$ samples. For the curves of evaluator $E$, we can observe that the RMSE loss raised for initial samples, then the RMSE value dropped significantly after few shots of data. For the curves toward responder $R$, the pattern is similar to the curves of $E$ (but in opposite way), the output quality score dropped for initial samples, while the score then bouncing back and achieving new highs with greater number of samples. Last but not least, as expected, the standard deviations lower along with the increasing number of training samples.

\end{document}